\documentclass{lmcs} %%% last changed 2014-08-20


\usepackage{algorithm2e}

\usepackage{hyperref}
\theoremstyle{plain} %\crefname{satz}{Satz}{S\"atze}
\def\eg{{\em e.g.}}

\begin{document}

\title[Segmentation-free CTC loss based OCR]{\textbf{Segmentation-free Connectionist Temporal Classification loss based OCR Model for Text Captcha Classification} }
\titlecomment{{\lsuper*}OPTIONAL comment concerning the title, \eg,
  if a variant or an extended abstract of the paper has appeared elsewhere.}
\thanks{thanks, optional.}	%optional

% affiliations are numbered automatically with a, b, c (see below)
% use the optional argument to indicate the affiliation(s) of each author
% omit the argument if there is only one author, or only one affiliation
\author[A.~Vaibhav]{Dr Vaibhav Khatavkar\lmcsorcid{0000-0002-1825-0097}}[a]
\author[B.~Makarand]{Dr Makarand Velankar\lmcsorcid{0000-0001-8946-0350}}[b]
\author[C.~Sneha]{Sneha Petkar 
\lmcsorcid{0009-0001-2783-1475}}
[c]

% affiliation 1 (automatically numbered a)
\address{VIT Bhopal University, Bhopal, Madhya Pradesh, India}	%optional
% write emails for all authors having that affiliation
\email{khatavkarvaibhav86@gmail.com}  %optional

% affiliation 2 (automatically numbered b)
\address{MKSSS's Cummins COE, Pune, Maharashtra, India}	%optional
\email{makarand.velankar@cumminscollege.com}  %optional
% affiliation 3 (automatically numbered b)
\address{GES's Sapat COE, Nashik, Maharashtra, India}	%optional
\email{snehapetkar10@gmail.com}
%% etc.

%% required for running head on odd and even pages, use suitable
%% abbreviations in case of long titles and many authors:

%%%%%%%%%%%%%%%%%%%%%%%%%%%%%%%%%%%%%%%%%%%%%%%%%%%%%%%%%%%%%%%%%%%%%%%%%%%

%% the abstract has to PRECEDE the command \maketitle:
%% be sure not to issue the \maketitle command twice!

\begin{abstract}
  \noindent Captcha are widely used to secure systems from automatic responses by distinguishing computer responses from human responses. Text, audio, video, picture picture-based Optical Character Recognition (OCR) are used for creating captcha. Text-based OCR captcha are the most often used captcha which faces issues namely, complex and distorted contents. There are attempts to build captcha detection and classification-based systems using machine learning and neural networks, which need to be tuned for accuracy. The existing systems face challenges in the recognition of distorted characters, handling variable-length captcha and finding sequential dependencies in captcha. In this work, we propose a segmentation-free OCR model for text captcha classification based on the connectionist temporal classification loss technique. The proposed model is trained and tested on a publicly available captcha dataset.  The proposed model gives 99.80\% character level accuracy, while 95\% word level accuracy. The accuracy of the proposed model is compared with the state-of-the-art models and proves to be effective.  
The variable length complex captcha can be thus processed with the segmentation-free connectionist temporal classification loss technique with dependencies which will be massively used in securing the software systems.  
\end{abstract}

\maketitle

%% start the paper here:
\section*{Introduction}\label{S:one}

In the digital world, large volumes of data are created and processed. Along with the large volume of data, the variety of data is another challenge. The data can be either text, image, audio, video, or a combination of text, image, audio, and video. The processing of combined data is usually performed by computer tasks defined as protocols and standards. Optical Character Recognition (OCR) is one such standard process used to read text data from the images. It has been more than a decade that researchers have developed various techniques in OCR \cite{Cite 0}.  Researchers in 
\cite{Cite 1}  have used a transformer model in the Neural Network for encoding and decoding OCR. Still today, OCR is applied in applications like car number plate detection \cite{App1}, cattle identification \cite{App2}, and text extraction from lecture slides \cite{App3}. 
Optical Character Recognition (OCR) is a field that has seen significant advancements, and various models and techniques have been developed to tackle the challenges posed by recognizing characters in captcha. Captcha, designed to be challenging for automated systems, often require specialized approaches.  Captcha is an acronym for Completely Automated Turing Test to Tell Computers and Humans Apart \cite{Ref1,Ref2,Ref3,Ref4}.  They are regularly used as a security measure for web services.  Captcha is an image embedded with text, audio or video data. \cite{C3}. The Text contents in the image are typically distorted. The text Captcha is generally a OCR, while audio and video Captcha are non OCR-based systems \cite{C2}. Though OCR-based Captcha is widely used, there are major issues in them \cite{C3}. These issues are :  \begin{itemize}
    \item \textbf{Distorted contents:} The captcha with text is distorted randomly. The distortion can make the text unreadable. 
\item \textbf{Complex contents:} The text captcha contains text with English digits and characters. For a user, the English language might be complex to read. The captcha are generated randomly, which may form a complex series of digits and texts, making the user difficult to understand and interpret. Sometimes these contents can be easily read by a computer program or an AI bot, making captcha breakable. 
\end{itemize}

Despite these issues, many applications still use captcha in their systems for human user verification.  The researchers in paper \cite{C1} review the captcha along with application and classification. OCR-based captcha are used for detection of captcha by computers.  The researchers in \cite{Ocrc1, Ocrc2, Ocrc3} have proposed computer-based solutions to detect OCR-based text captcha using methods like Deep Neural Networks, Machine Learning and CNN [ref5, ref6, ref7]. 
The motivation behind developing an OCR (Optical Character Recognition) model specifically designed for reading captcha stems from several important factors related to online security, user authentication, and the evolving landscape of automated threats. There is a need to provide sophisticated a method to provide security to the text-based captcha to increase its robustness against the attacks.  To be effective, captcha need to be sufficiently complex to prevent automated systems from easily deciphering and responding to them. This complexity often involves distorted characters, varying fonts, and other obfuscation techniques that can be challenging for traditional OCR systems. Also, conventional OCR models, which are designed for general text recognition, may struggle to accurately interpret characters in captcha images. 
Thus, to recognize the distorted characters efficiently, handle variable length of captcha and resolve sequential dependencies a novel OCR based captcha detection approach is proposed in this paper.

\section{Related Work}
Deep learning techniques are widely used in feature extraction from the images and has a variety of applications in image restoration \cite{Ref6, Ref7, Ref8} and object detection \cite{Ref9, Ref10, Ref11}. The high computational power of these techniques has made it a good choice to build a captcha detection system. Despite the high computational power, these techniques suffer due to weak feature extraction processes and noisy input images. Thus, new techniques in deep learning are to be proposed for captcha detection. 
The segmentation-based and segmentation-free are two widely used techniques for text-based captcha detection. The segmentation-based systems use either segmentation or character recognition algorithm \cite{Ref12, Ref13}.  The segmentation algorithm is a two-step process with the segmentation of image as first followed by character recognition Algorithm \ref{Algo:capatca_identification}. 
Despite being a simple two-step process, the segmentation-based captcha detection systems have low efficiency and low accuracy as well.  To improve efficiency and accuracy, segmentation-free captcha detection algorithms are used. The segmentation-free algorithms recognize and classify the characters directly, i.e. without segmenting the image into individual characters. 

\begin{algorithm} \label{Algo:capatca_identification}    
Algorithm 1: Segmentation Algorithm \cite{Ref12, Ref13} \\ 
Step 1: Segmentation of captcha into individual characters using segmentation module \\
Step 2: Individual character recognition using a character recognition module. 
    
\end{algorithm}

 To train the deep learning models for captcha recognition with necessary image features, large training datasets are required to increase the efficiency of the algorithm. The segmentation-free models, though widely used, have complex architectures. They also require relatively large storage resources, which depend on the number of characters in the image.
Apart from this, Captcha often includes intentional distortions that make them inherently different from standard text, thus requiring a specialized approach.
captcha typically involve characters that are intentionally distorted, rotated, or presented in a way that is challenging for conventional OCR models. The OCR model needs to be robust enough to handle these distortions and accurately recognize the characters.  captcha can have varying lengths, which means the OCR model should be capable of handling sequences of characters with different lengths. This requirement makes traditional fixed-size input OCR models less suitable, necessitating the use of techniques like Connectionist Temporal Classification (CTC) loss.
Captcha images typically contain characters arranged in a sequence, where the correct interpretation of one character depends on the context provided by the surrounding characters. Recurrent Neural Networks (RNNs) are particularly effective in capturing these sequential dependencies. Thus, to recognize the distorted characters efficiently, handle variable length of captcha and resolve sequential dependencies a novel OCR based captcha detection approach is proposed in this paper.  The template based OCR techniques are susceptible to variations in font, size, and orientation \cite{Ref17, Ref18, Ref19}. They are also ineffective for distorted or obfuscated characters in captcha. The feature-based approaches have limited adaptability to highly distorted characters and are sensitive to variations in scale and orientation. In the development of OCR, deep learning models like CNN \cite{Ref14,Ref15}, and RNN have played a pivotal role. The CNN when used in OCR provided an excellent feature extraction technique for image-based tasks. The OCR model with CNN can capture hierarchical features, useful for recognizing patterns. RNN are proven to be effective in capturing sequential dependencies in OCR. OCR with RNN are found suitable for recognizing characters in a specific order.  CRNNs integrate CNNs and RNNs to capture both spatial and sequential information. They are effective for recognizing characters in variable-length sequences.  
CNN-based segmentation-free models for captcha detection are proposed by various researchers in \cite{Ref20, Ref21, Ref22}. Some segmentation-free models combine CNN and attention-based recurrent neural network (RNN) for captcha detection \cite{Ref23, Ref24, Ref25}. 

The Connectionist Temporal Classification (CTC) Loss for Variable-Length Sequences is another model for OCR development. This model handles variable-length sequences without requiring explicit alignment. It is also useful for training models to handle the dynamic nature of captcha lengths.  Therefore, in this paper, we propose a Connectionist Temporal Classification loss based OCR Model for Text Captcha Classification.

\section{Proposed System}

\begin{figure}[ht]
    \centering
   \includegraphics[scale=0.55]{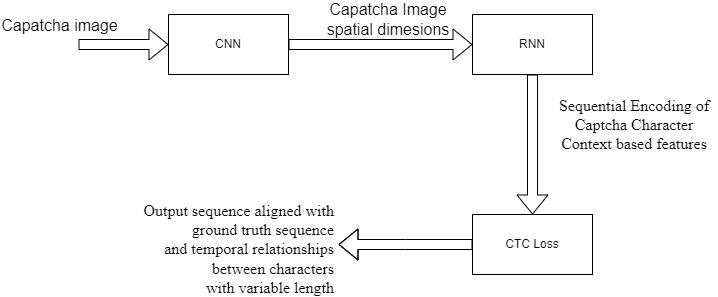}
    \caption{Proposed Model}
    \label{fig:proposed_model}
\end{figure}

The architecture of an OCR (Optical Character Recognition) model that incorporates Convolutional Neural Networks (CNNs), Recurrent Neural Networks (RNNs), and Connectionist Temporal Classification (CTC) loss is designed to handle the challenges posed by reading captcha and is shown in Figure \ref{fig:proposed_model}. 

The components of the proposed model with their functionalities are stated below: 
\begin{itemize}
    \item 
\textbf{Convolutional Neural Networks (CNNs):} The Convolutional Neural Network (CNN) layers used for feature extraction from captcha images play a crucial role in capturing relevant spatial and hierarchical features. The following outline provides a common structure for the CNN layers used in OCR models for captcha recognition. 
\item 
\textbf{Recurrent Neural Networks (RNNs):} They play a crucial role in OCR models for captcha recognition by capturing sequential dependencies in the features extracted by Convolutional Neural Networks (CNNs). captcha often involve characters arranged in a specific order, where the recognition of one character depends on the context provided by the surrounding characters. 
The output from the last dense layer of the CNN, representing the features extracted from the spatial dimensions of the captcha image, is used as input to the RNN. The RNNS then process the RNNs process the input sequence sequentially, processing one feature at a time while maintaining hidden states that capture information from previous steps. The sequential processing capability of RNNs makes them suitable for tasks where the order of input matters, as is the case in reading captcha.  The hidden states of the RNN at each time step capture relevant information about the sequential context. These hidden states are updated and propagated through the network as each character is processed. The output sequence from the RNN represents the sequential encoding of features based on the context of the characters in the captcha.  Each element of the output sequence corresponds to the hidden state at a specific time step.
\item 
\textbf{Connectionist Temporal Classification (CTC) Loss:} It is a technique used in machine learning, particularly in the context of sequence-to-sequence tasks, where the input and output sequences can have variable lengths. It is commonly used in tasks like speech recognition, handwriting recognition, and machine translation. \\
In scenarios where the input and output sequences have different lengths or are not aligned one-to-one, traditional loss functions like mean squared error or cross-entropy may not be suitable. This misalignment issue arises in tasks where the length of the input sequence does not necessarily correspond to the length of the target sequence.
\\CTC loss addresses this problem by allowing the model to learn alignments between the input and output sequences automatically. It was initially developed for speech recognition but has since been applied to various sequence-to-sequence tasks. \\ 
The Output sequence of RNN is then passed to the CTC Loss layer, which allows the model to produce a new output sequence that aligns with the ground truth sequence. The CTC layer facilitates the learning of the temporal relationships between characters and helps handle variable-length sequences. During training, the OCR model is optimized using the CTC loss, which calculates the difference between the predicted sequence and the ground truth sequence. The CTC loss guides the model to learn the alignment between the input sequence and the corresponding output sequence, handling the variability in captcha lengths. 
\end{itemize}
During inference, CTC decoding algorithms are applied to convert the model's output sequence into the final recognized text.  Decoding methods handle variable-length sequences and provide the predicted text by considering the probabilities of different character alignments.
\section{Experiments and results}
\begin{figure}[ht]
    \centering
    \includegraphics[scale=1.5]{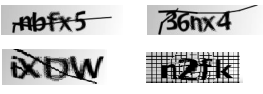}
    \caption{Sample data}
    \label{fig:sample_data}
\end{figure}

The choice and quality of the dataset play a crucial role in the training and evaluation of an OCR model for a captcha. A well-constructed dataset that accurately represents the challenges posed by captcha can significantly impact the model's performance.  We used the text captcha image dataset which is publicly available on Kaggle  \cite{Refdata}. The dataset contains 1040 samples with Text Captcha Images and the label for each sample is the name of the file which is the text present inside the captcha, The captcha in the dataset vary in terms of distortion, font type, background noise, and character complexity.  The character set in the captcha reflects the expected diversity of characters in real-world scenarios. Alphanumeric characters (both uppercase and lowercase) and special symbols commonly used in captcha are included. There are 19 unique characters found in the dataset namely : ['2', '3', '4', '5', '6', '7', '8', 'b', 'c', 'd', 'e', 'f', 'g', 'm', 'n', 'p', 'w', 'x', 'y']. To simulate real-world scenarios, the dataset includes a captcha with variable-length sequences. captcha with different numbers of characters challenges the model to handle sequences of varying lengths. The sample data from the dataset is shown in Figure \ref{fig:sample_data}.  

\begin{figure}[ht]
    \centering
   \includegraphics[scale=0.4]{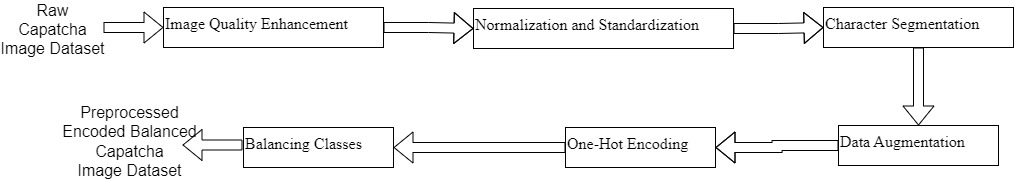}
    \caption{Data Pre-processing Steps}
    \label{fig:data_preprocessing}
\end{figure}

\begin{figure}[ht]
    \centering
    \includegraphics[scale=0.35]{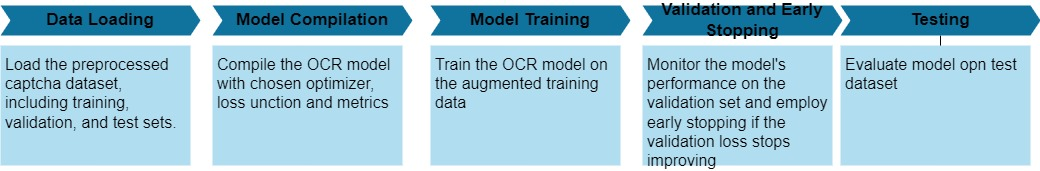}
    \caption{Training of model}
    \label{fig:training}
\end{figure}

The data pre-processing involves tasks as shown in Figure \ref{fig:data_preprocessing}. Captcha images may vary in quality. Pre-processing steps such as contrast adjustment, normalization, and noise reduction enhance the overall image quality. Image quality enhancement also ensures uniformity in image quality, helps the model generalize better. Normalizing pixel values to a specific range (e.g., [0, 1]) ensures consistent input to the model. While the Standardization of pixel values (mean subtraction and scaling) may be applied to improve convergence during training. If the dataset includes captcha with characters merged or overlapping, a preprocessing step for character segmentation is crucial.  Proper segmentation ensures that individual characters are presented to the OCR model for accurate recognition.  Data Augmentation techniques such as rotation, scaling, translation, and flipping are applied to artificially increase the diversity of the dataset. Augmentation helps the model generalize better and become more robust to variations in real-world captcha images. The ground truth labels, representing the characters in the captcha, undergo one-hot encoding.  This encoding converts categorical labels into a binary matrix, facilitating the training of the model with categorical cross-entropy loss. The Balancing Classes step ensures that the dataset has a balanced representation of different captcha types and character classes. Imbalanced datasets can lead to biased model training, affecting the model's performance on under-represented classes.
\\ After getting the preprocessed encoded balanced dataset, it is divided into training, validation and testing dataset.  The majority of the dataset is allocated for training the OCR model. Typically, around 70-80\% of the dataset is used for training to allow the model to learn the underlying patterns in the captcha.  A smaller portion (e.g., 10-15\%) is set aside for model validation during training. The validation set helps in monitoring the model's performance on unseen data and prevents overfitting. A separate test set, not seen by the model during training, is reserved for evaluating the model's generalization. The test set contains a diverse set of captcha, ensuring a comprehensive assessment of the model's performance. \\
The training process for an OCR model designed for reading captcha involves several critical components, each of which contributes to the model's performance and generalization ability. The training process, includes training hyper-parameters, optimizer choice, and data augmentation techniques (image — rotation, flipping, translation, zooming, shear, brightness adjustments, and contrast variations ). The training Steps are as shown in figure \ref{fig:training}. 

The model has been implemented in Python with libraries like TensorFlow, Keras, OpenCV, NumPy, and nltk. The datasets available in TensorFlow\_datasets  are used to load and preprocess the dataset during the training and validation phase. The training visualization is down with the Tensorboard. These tools provide visualization and monitoring of the training process, including metrics, loss curves, and model architecture visualization.
\begin{figure}[ht]
    \centering
    \includegraphics[scale=0.35]{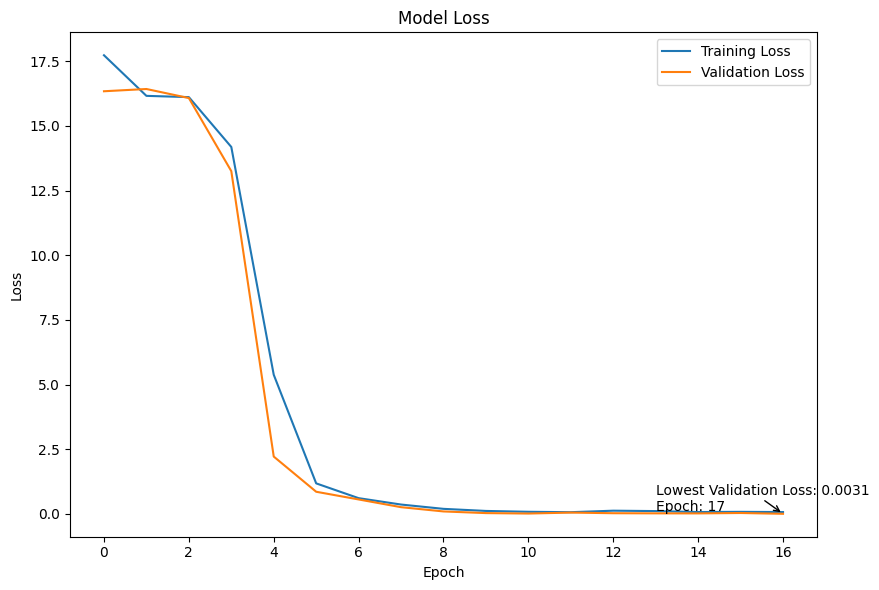}
    \caption{Loss Curve}
    \label{fig:loss_graph}
\end{figure}

The training and validation CTC loss over epochs is shown in the graph (Figure \ref{fig:loss_graph}). A decreasing loss indicates that the model is learning and converging.  Analyzing the results of an OCR project involves evaluating the performance of the trained model on various metrics, understanding the challenges faced during the project, and identifying potential improvements for the model.  To evaluate the model, we use two level accuracy metrics, namely, character level and word level accuracy. Character level accuracy gives the accuracy of correctly predicted characters by the model, which helps to understand how well the model recognizes individual characters. Word level accuracy is the accuracy of correctly predicted words by the model. Word level accuracy is a more practical evaluation metric since OCR often recognizes words.
\begin{table}[ht] 
\resizebox{\textwidth}{!}{ 

    \centering
    \begin{tabular}{|c|c|c|}
      \hline Type   & Proposed Model  Accuracy (\%)  & State-of-art Model  Accuracy (\%)\\ \hline
    Character Level  Accuracy     & 99.80  &  98.94  %\cite{Ref}, 
    \cite{C} ,  99.70 \cite{A}, 51.1 \cite{Compref1}, 51.6 \cite{Compref2}  
 \\ \hline   Word Level   Accuracy & 
95 & 
94 \cite{B}
\\ \hline
    \end{tabular}
    }
    \caption{Evaluation metric}
    \label{tab:evaluation_metric}
\end{table}

The proposed model gives 99.80\% character level accuracy, while 95\% word level accuracy. The accuracy of the proposed model is compared with the state-of-the-art models and proves to be effective.  The system has limitations due to factors like, the limited unrepresentative training data, noisy and distorted images, variable length sequences (though padding is used in RNN), hyper-parameter values (learning rate, batch size, etc.)  and model complexity.  

\section{Conclusion}
The OCR model demonstrated proficiency in recognizing captcha, showcasing its ability to decipher distorted, variable-length, and complex patterns commonly used in security measures on websites. The combination of Convolutional Neural Networks (CNNs) for spatial feature extraction and Recurrent Neural Networks (RNNs) for capturing sequential dependencies proved highly effective in handling the unique challenges posed by captcha. \\ 
The use of Connectionist Temporal Classification (CTC) loss facilitated training on variable-length sequences, allowing the model to adapt to different captcha lengths without requiring explicit alignment information. Data augmentation techniques, including rotation, flipping, translation, and zooming, significantly enhanced the model's robustness by exposing it to a diverse range of captcha variations during training. Careful tuning of hyperparameters, such as learning rate, batch size, and network architecture, played a pivotal role in achieving optimal model performance during training.  The OCR model's ability to decipher captcha contributes directly to addressing the need for an effective solution against automated systems attempting to bypass security measures. The proposed system is developed which can handle distorted text, complex character combinations and detect dependencies in character sequences.  The model's successful recognition of diverse captcha suggests promising generalization to unseen variations, contributing to its potential application in real-world scenarios. The findings provide a strong foundation for further development, including exploration of attention mechanisms, ensemble methods, and real-time application adaptation. \\ 
The possible future research and improvements to enhance the OCR model for reading captcha is to investigate the use of GANs to generate synthetic captcha images for augmenting the training dataset. GANs can create realistic variations, providing additional diversity to improve the model's generalization. Explore transfer learning by leveraging pre-trained models on large image datasets. Fine-tuning the OCR model on captcha data after pre-training on general image datasets may improve convergence and performance.
\\ Analyzing OCR model results is an iterative process that involves a combination of quantitative metrics, qualitative assessments, and a deep understanding of the challenges faced. Continuous refinement based on insights gained from analysis is key to building a robust OCR system. Regularly revisiting and updating the model based on new data and user feedback is essential for long-term success.\\ 
Explore multimodal approaches by integrating additional information, such as audio cues or contextual information, to enhance captcha recognition accuracy, especially in scenarios where visual information alone may be insufficient. By pursuing these research directions, the OCR model for reading captcha can be enhanced in terms of accuracy, robustness, and adaptability to diverse challenges in real-world applications.

\end{document}